\documentclass{ieeeaccess}
\usepackage{cite}
\usepackage{amsmath,amssymb,amsfonts}
\usepackage{algorithmic}
\usepackage{graphicx}
\usepackage{textcomp}
\usepackage{multirow}
\usepackage{soul}
\usepackage{caption}

\def\BibTeX{{\rm B\kern-.05em{\sc i\kern-.025em b}\kern-.08em
    T\kern-.1667em\lower.7ex\hbox{E}\kern-.125emX}}

\begin{document} 
\history{}
\doi{}

\title{Alzheimer’s Magnetic Resonance Imaging Classification Using Deep and Meta-Learning Models}
\author{\uppercase{Nida Nasir}\authorrefmark{1*}, \uppercase{Muneeb Ahmed}\authorrefmark{2},
\uppercase{Neda Afreen} \authorrefmark{2},
\uppercase{Mustafa Sameer} \authorrefmark{3}}
\address[1]{University of Gustave Eiffel, Lille, France. }
\address[2]{Department of Computer Engineering, Jamia Millia Islamia University, New Delhi, India }
\address[3]{Department of Electronics and Communication Engineering, National Institute of Technology Patna, India}
\markboth
{Nasir \headeretal: Alzheimer’s Magnetic Resonance Imaging Classification Using Deep
and Meta-Learning Models}
{Nasir \headeretal: Alzheimer’s Magnetic Resonance Imaging Classification Using Deep
and Meta-Learning Models}

\corresp{Corresponding author: Nida Nasir (e-mail: nida.nasir@univ-eiffel.fr).}

\begin{abstract}
Deep learning, a cutting-edge machine learning approach, outperforms traditional machine learning in identifying intricate structures in complex high-dimensional data, particularly in the domain of healthcare. This study focuses on classifying Magnetic Resonance Imaging (MRI) data for Alzheimer’s disease (AD) by leveraging deep learning techniques characterized by state-of-the-art CNNs. Brain imaging techniques such as MRI have enabled the measurement of pathophysiological brain changes related to Alzheimer's disease. Alzheimer's disease is the leading cause of dementia in the elderly, and it is an irreversible brain illness that causes gradual cognitive function disorder. In this paper, we train some benchmark deep models individually for the approach of the solution and later use an ensembling approach to combine the effect of multiple CNNs towards the observation of higher recall and accuracy. Here, the model's effectiveness is evaluated using various methods, including stacking, majority voting, and the combination of models with high recall values. The majority voting performs better than the alternative modelling approach as the majority voting approach typically reduces the variance in the predictions. We report a test accuracy of 90\% with a precision score of 0.90 and a recall score of 0.89 in our proposed approach. In future, this study can be extended to incorporate other types of medical data, including signals, images, and other data. The same or alternative datasets can be used with additional classifiers, neural networks, and AI techniques to enhance Alzheimer's detection.

\end{abstract}


\titlepgskip=-15pt

\maketitle
\section{Introduction}
Alzheimer's Disease (AD) is the third most prevalent illness after cancer and cardiovascular illnesses, and the sixth major cause of mortality \cite{alzheimer20162016}. According to estimates, there are currently 44.4 million dementia patients worldwide, and that figure is expected to rise to 75.6 million by 2030 and 135.5 million by 2050 \cite{vradenburg2015pivotal}. In the past century, AD diagnosis was based on clinical criteria (laboratory tests, neuropsychological tests, and imaging findings). These criteria have an accuracy of 85\% and do not provide a conclusive diagnosis; this can only be done through post-mortem examination. Therefore, machine learning (ML) has been employed in the last decade to discover Magnetic Resonance Imaging (MRI) biomarkers for Alzheimer’s disease. Many ML technologies are now being used to improve the identification and forecasting of Alzheimer’s disease. The application of deep learning to the early detection and automated classification of AD has recently received significant attention, thanks to rapid advancements in neuroimaging techniques that have generated a lot of multimodal neuroimaging data. Deep learning models have recently been shown to be the best for neuroimaging data related to Alzheimer's disease. They efficiently and effectively represent the information by retrieving data from images. Health specialists can use ML and AI to diagnose and predict disease risks more accurately and quickly, allowing them to prevent diseases in time.

In the proposed study, we try to utilize the efficiency of different models like ResNet, Squeezenet, VGG, Inception V3 and Mobilenet. In medical science data, the best performance is observed based on its recall value, as it is very important to say every positive outcome is positive in this type of problem. The MRI dataset of ADNI was in a nifty format. First, it was converted into video format and then split into images of around 150 samples. The first and last few samples, around 30 to 40, were redundant, having no significant information. Most of the MRI data is redundant, so it is removed before modeling, and the most prominent samples are taken by considering sample's entropy. The samples are taken in three ways: first, only one sample has the maximum entropy or contains maximum information. The second takes the top 50 samples having the highest entropy value, and last all samples combined. The best results come from the top 50 samples compared to the other two sample types because those samples contain most of the information in a limited subset of frames. The model's performance is measured here through different approaches: by combining models with high recall value, through stacking, and by majority voting. We find that the majority voting method performs better than other modelling strategies as the majority voting method tends to minimize the variance in the predictions by taking mode over the predictions of the three other models.

In this paper, we explore the solution to the problem in coherence to the following contributions:
\begin{itemize}
    \item Comparing the performance of benchmark CNNs in classifying the MRI scans. 
    \item Using ensemble approach over multiple CNNs towards achieving better performance metrics.
    \item Experimentally substantiating a correlation between information in the data characterized by entropy and the performance of the deep neural networks used.
    \item Achieving a higher recall score using a majority voting scheme. 
     
\end{itemize}

\section{Literature review}
Researchers are attempting to establish a reliable method of diagnosing AD based on multiple different classification methods. This section will focus on some of the study publications that are connected to various classification strategies that are widely used for Alzheimer’s disease. Oppedal et al. [1] proposed a comprehensive AD categorization approach. The authors have carried out an operation to create white matter lesions (WML) maps from the input pictures as the first phase in the process. In the second stage, the scientists extracted texture information from the maps by using a local binary pattern-based technique. In the end, all of the pertinent aspects of the texture are extracted, and a Random Forest-based classification is carried out in order to classify AD cases. A different strategy to categorization that is Random Forest-based was proposed by Ardekani et al. [2]. It was determined that the region of interest was the hippocampus in the brain, and a process called segmentation was carried out to separate it. The classification is carried out using Random Forest method that is made up of 5000 different decision trees. In the research literature [3], there is a classification approach for AD that is based on random forests. In order to do the necessary preprocessing and segmentation on the input photos, the authors relied on FreeSurfer toolbox. An approach that is based on the Gini impurity index is utilized in order to extract the features that are the most relevant. The final classification is carried out with the help of an algorithm known as random forest, in which decision trees are constructed based on a variety of bootstrap samples.

Many researchers opted to choose KNN and SVM for their classification work. Being able to learn from the environment and enhance its performance with each iterations is one of ANN's key advantages. KNN is simple, both in terms of its explanation and its application. It is effective in solving problems involving multiple classes. A DNN-based methodology was developed for the categorization of AD by Lu et al. [4]. After performing the gray matter segmentation procedure in the beginning, the authors retrieved patch-wise information in order to train a multimodal and multiscale deep neural network (MMDNN). The created network is composed of two parts: the first portion takes into consideration six separate DNNs for each modality, and the second part is utilized to combine the information that was mined from the six DNNs. The authors Kamathe et al. [5] developed an automatic categorization of AD that was based on a robust and optimized feature set. Using a technique known as gray level co-occurrence matrix (GLCM), the authors' primary objective is to isolate the characteristics that are the most important. The authors have made use of an Adaboost-based method in order to improve the overall performance accuracy. The KNN classifier is used, after the extracted feature sets have been improved, to perform classification in accordance with the results. Tufail et al. [6] proposed developing a mechanism to categorize Alzheimer's disease using several different classifiers. They did this by making use of structural MRI. The authors have chosen a strategy that is based on independent component analysis (ICA) in order to extract the features that are capable of discrimination. After acquiring the extracted features, the authors used three distinct classifiers, namely SVM, ANN, and KNN, and compared the output outcomes of each classifier. According to what the authors assert, ICA+KNN produces results that are compelling.

Transfer learning approach is widely popular in medical research as it involves training networks on a specific dataset (even an irrelevant one) as initialization, and then retraining them on a new dataset by just fine-tuning the networks. This is done in order to maximize the likelihood of the networks producing accurate results on the new dataset. Three 2D CNNs with two convolutional layers were trained by Aderghal et al. [7] using only three slices in the hippocampal region's center in some MRI images. Instead of starting from scratch while only having a small number of DTI pictures, they used transfer learning to apply models that had already been trained on MRI images to the target dataset. Finally, they integrated all the networks and used a majority voting method to get a result. In a 2D convolutional neural network, the network is incapable of separately examine the relationships between 2D picture slices inside an MRI volume. Amir et al.[8] suggest using a recurrent neural network following a convolutional neural network to comprehend the relationship between sequences of images for each individual and make a decision based on all input slices rather than just one slice to overcome this issue. According to their findings, the accuracy of the entire system can be increased by training the recurrent neural network using features taken from a convolutional neural network.

Ensemble learning utilizes a group of decision-making systems that make use of a range of approaches in order to combine classifiers with an improved prediction of data. This is accomplished through the use of "ensembles." In fields relating to healthcare, ensemble learning has been utilized to boost the accuracy of AD prediction by employing the clinical expertise of ordinary clinicians. This has been done in order to improve patient outcomes. It is possible to use it in primary care settings, which typically have limited access to specialty care providers.
All literature works are summarized in Table 1.

\begin{table*}[ht]
\caption{Brief description of work in literature review.}
\centering
 \resizebox{18cm}{!}{%
\begin{tabular}{llllllllllll} \hline \hline
\textbf{Study} & \multicolumn{5}{l}{\textbf{Subjects}} & \textbf{Classification Technique} & \textbf{Database} & \multicolumn{4}{l}{\textbf{Accuracy}}        \\
               & AD   & MCI   & pMCI   & aMCI   & CN   &                             &                    & AD/CN & MCI/CN & AD/MCI & aMCI/pMCI \\ \hline
  \cite{casanova2011high}             &    49  &    -   &  -      &     -   &   49   &              LSR               &         ADNI           &    85.75   &   -     &     -   &       -    \\
      \cite{termenon2012two}         &      49&       -&        -&        -&      49&                             RVM, SVM&                    ADNI&       83.0&        -&        -&           -\\
    \cite{adaszewski2013early}   &      108&-       142&        61&        137&      SVM&                             ADNI&                    -&       -&        -&        62.0&     \\   
     \cite{plant2010automated}   &  32    &-&       9&        15&        18&      SVM, BS, VFI&                             CPLMU&                    92.0&       -&        -&        75.0     \\    
       \cite{moller2016alzheimer}   &    84  &       -&        -&        -&      94&                             SVM&                    ACVUMC, ACEUMC&       88.0&        -&        -&     -\\   
    \cite{liu2013locally}    &      86&       -&        97&        93&      137&                             SVM, LDA&                    ADNI&       90.0&        -&        -&     68.0\\     
       \cite{salvatore2015magnetic}   &      137&       -&        76&        134&      162&                             SVM&                    ADNI&       76.0&        72.0&        -&     66.0\\   
     \cite{beheshti2015probability}   &      130&       -&        -&        -&      130&                             SVM&                    ADNI&       86.65&        -&        -&     -\\     
      \cite{min2014multi}    &      97&       -&        117&        117&      128&                             SVM&                    ADNI&       91.64&        -&        -&     72.41\\   
   \cite{liu2015view}     &      97&       -&        117&        117&      128&                             SVM&                    adni&       92.51&        -&        -&     78.88\\     
       \cite{sorensen2016early}   &      101&       233&        93&        140&      169&                             SVM&                    ADNI&       91.20&        76.40&        -&     74.20\\   
       \cite{tang2015baseline} &   175   &       -&        135&        87&      210&                    LDA & ADNI                    &       -&        -&    -&    74.77     \\     
      \cite{challis2015gaussian}    &      27&       50&        -&        -&      39&                             GP-LR&                    Internal&       -&        75.0&        97.0&     -\\   
     \cite{jie2013integration}   &     - &       12&        -&        -&      25&                             Multi-kernel SVM&                    BIAC&       -&        97.0&        -&     -\\     
     \cite{khazaee2015identifying}     &      -&       20&        -&        -&      20&                             SVM&                    ADNI&       100&        -&        -&     -\\   
      \cite{nir2015diffusion}  &      37&       113&        -&        -&      50&                             SVM&                    ADNI&       80.6&        68.3&        -&     -\\     
       \cite{prasad2015brain}   &      38&       -&        38&        74&      50&                             SVM&                    ADNI&       78.2&        62.8&        -&     63.4\\  
       \cite{dyrba2013robust}     &      137&       -&        -&        -&      143&                             SVM&                    EDSD&       83.0&        -&        -&     -\\  
\cite{dyrba2015predicting}              &      -&       -&        35&        35&      25&                             SVM&  EDSD                  &       -&        77&     -   & 68    \\  
          \cite{wee2011enriched}      &      -&       10&        -&        -&      17&                             SVM&                    BIAC&       -&        88.90&        -&     -\\  
                  \cite{korolev2016predicting}&      -&       -&        139&        120&      -&                            Probabilistic multi-kernel &                    &       -&        -&        -&     80.0\\  
             \cite{yu2016graph}       &      50&       97&        -&        -&      52&      Multi-task learning&                       ADNI&                    92.60&       80.0&        -&        -     \\  
\cite{li2018alzheimer}                      &  199    &       403&        -&        -&      229&                             DenseNet&                    ADNI&       89.8&        74.0&        -&     -\\  
      \cite{liu2020new}                  &23      &       -&        -&        -&      13&                             LogisticregressionCV&      VBSD              &       84.4&        -&        -&     -\\  
            \cite{battineni2021deep}            &   489   &       -&        -&        -&      609&                             ConvNet&                    OASIS&       80.0&        -&        -&     -\\ \hline \hline
\end{tabular} %
}
\label{literature}
\end{table*}

\begin{table}[]
\caption{}
\begin{tabular}{p{0.09\textwidth}ccc} \hline \hline
\multicolumn{1}{c}{\multirow{2}{*}{\textbf{Authors \& Articles}}} & \multicolumn{3}{c}{\textbf{Performance}}                                                                                 \\
\multicolumn{1}{c}{}                                     & \multicolumn{1}{l}{\textbf{Accuracy (\%)}} & \multicolumn{1}{l}\textbf{{\textbf{Sensitivity (\%)}}} & \multicolumn{1}{l}{\textbf{Specificity (\%)}} \\\hline
Oppedal et al. \cite{oppedal2015classifying} (RF)                          & 87                                & -                                    & -                                    \\
Ardekani et al. \cite{ardekani2017prediction} (RF)                   & 82.3                              & 86                                   & 78.2                                 \\
Kim et al. \cite{kim2021development} (RF)      & 85.03                             & -                                    & -                                    \\
Lu et al. \cite{lu2018multimodal} (ANN)                                & 82.4                              & 94.23                                & 86.3                                 \\
Kamathe et al. \cite{kamathe2018robust} (KNN)                                           & 87                                & -                                    & -                                    \\
Tufail et al. \cite{tufail2012automatic} (KNN)                                           & 68.06                             & -                                    & -                                    \\
Aderghal et al. \cite{aderghal2018classification} (TL)                                        & 80                                & 66.7                                 & 60                                   \\
Amir et al. \cite{ebrahimi2019transfer} (TL)                                         & 71.88                             & -                                    & -                 \\ \hline \hline                  
\end{tabular}
\label{Tab:}
\end{table}

\section{Methodology}
This section discusses the dataset used, the preprocessing steps, and the deep learning models used along with the ensemble stacking.

\begin{figure*}[ht!]
    \centering
    \includegraphics[width=\linewidth]{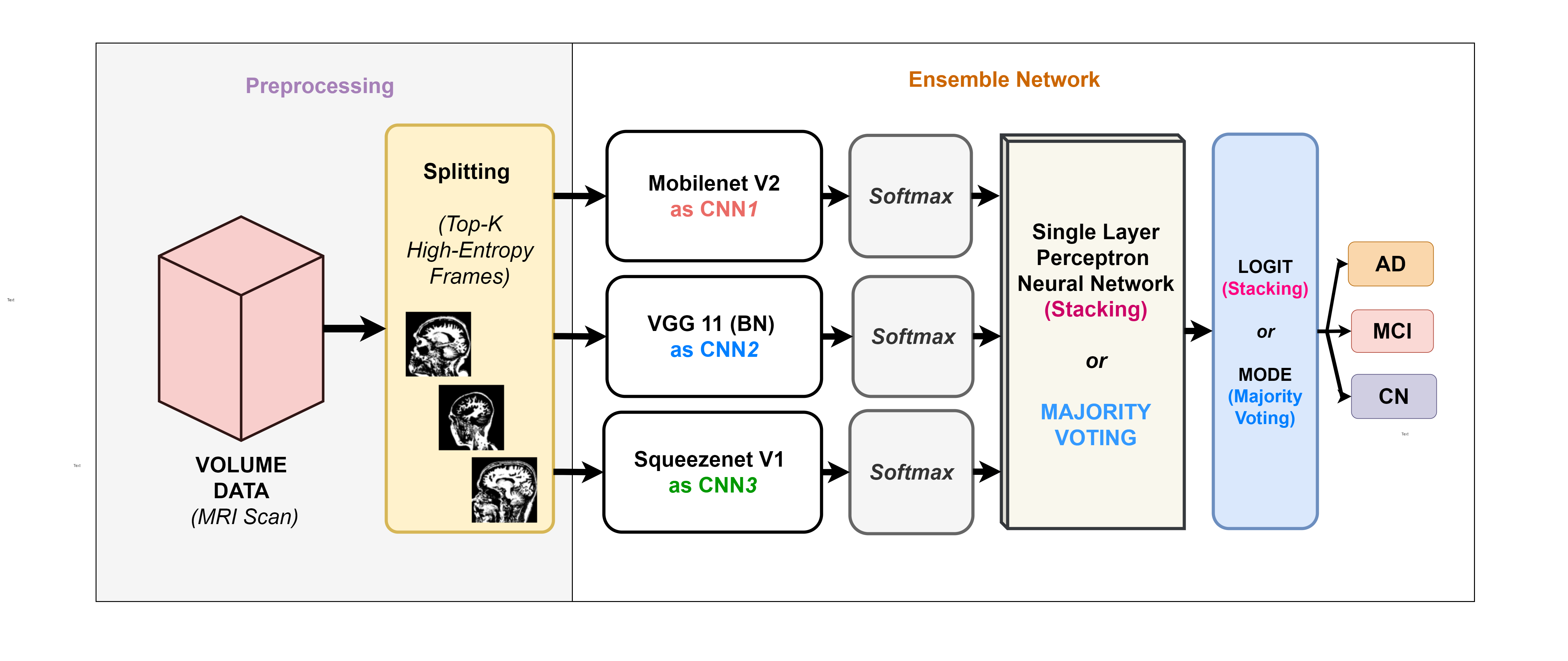}
    \caption{Workflow of the proposed system.}
    \label{fig:workflow}
\end{figure*}

The workflow of the proposed methodology is two-fold viz.,
\begin{enumerate}
    \item Procuring Dataset and Preprocessing
    \item Modelling/Training
 \end{enumerate}
Owing to the nature of volume-data in the training set, it is imperative to transform the input into an interpretable format with respect to the training algorithms. The second step involves various training algorithms that can model the features in the data.
Fig. \ref{fig:workflow} illustrates the end-to-end workflow of the proposed system.

\subsection{Dataset}
\begin{figure*}[hbt!]
    \centering
    \includegraphics [scale=1]{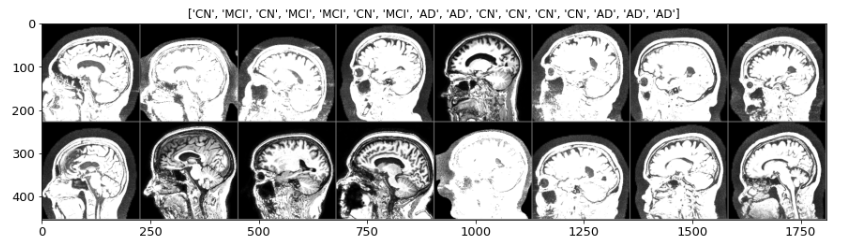}
    \caption{Sample population of Brain MRI Scans from ADNI dataset.}
    \label{fig:BrainMRIsamples}
\end{figure*}

The Alzheimer's Disease Neuroimaging Initiative (ADNI) repository provided the MR images of the brain, which are used in this study \cite{adni}. The ADNI was established in 2003 as a 60 million dollar, 5-year public-private partnership by the Food and Drug Administration (FDA), the National Institute of Biomedical Imaging and Bioengineering (NIBIB), the National Institute on Aging (NIA), private pharmaceutical companies, and non-profit organizations. The ADNI's main purpose was to see if serial MRI, PET, other biological markers, and clinical and neuropsychological assessments could be used to track the evolution of mild cognitive impairment (MCI) and preclinical Alzheimer's disease (AD). The goal of identifying potent and selective signs of quite initial Alzheimer's disease development is to help researchers and doctors to create drug therapies and assess their effectiveness while also reducing the cost and time of clinical studies. Michael W. Weiner, M.D., of the VA Medical Center and the University of California, San Francisco, is the initiative's primary investigator. Many co-investigators from a variety of educational institutions and corporate corporations worked together to create ADNI. The study participants came from more than 50 locations across the United States and Canada and submitted informed written consent for imaging and genomic samples collection, as well as answered questionnaires permitted by every contributing site's Institutional Review Board (IRB).
We evaluate T1-weighted MR scanning datasets from basic of 352 participants, including 77 Alzheimer's disease (AD), 145 moderate cognitive impairments (MCI), and 129 healthy controls (NC). Table 2 shows the specifics of the subjects used. The volumetric 3D MPRAGE with 1.25x1.25 mm$^2$ in-plane spatial resolution and 1.2 mm thick sagittal slices is used in ADNI to capture T1-weighted MR images sagittally. The scans were taken with 3T scanners. The acquisition information is concerning to the vendors of this dataset. The MR images are processed in the following manner: First, the intensity inhomogeneity in the scans is corrected with a non-parametric non-uniform intensity normalization (N3) technique. This is followed by skull-stripping and eliminating the cerebellum region in the images. In the last step, linear registration correlates the MR scans to a template. A sample distribution of the time-splitted scans is shown in Fig. \ref{fig:BrainMRIsamples}. Different versions of the dataset from the vendors exist based on the span of scans and the technology used. The dataset considered in this study is ADNI 3Yr 3T. Table \ref{tab:my_label} shows the distribution of data in the considered dataset.

\begin{table}[]
    \centering
    \caption{Distribution characteristics of \emph{ADNI 3Yr 3T dataset}.}
    \begin{tabular}{llll}
    \hline \hline
    \textbf{Diagnosis}   &  \textbf{Number} & \textbf{Age} & \textbf{Gender(M/F)} \\ \hline
    AD   &  77& 57-90 & 32/45 \\
    MCI  &  145  & 55-90  &  111/34  \\
    CN  &  129 & 70-88  &  51/78 \\ \hline \hline
    \end{tabular}
    \label{tab:my_label}
\end{table}

\subsection{Preprocessing}
An MRI scan is characterized as spatially varying time-series image data. Every snapshot records the part of the body under investigation at a preset frame rate. Not all frames contain sufficient and complete information. We observed that only a subset of such frames represents a significant portion of the entropy, which is justified by our experiments mentioned later in this paper. 
Initially, we extracted the frames from each recording instance of a certain subject and divided the dataset into three types of samples viz., (i) only one sample with maximum entropy, (ii) top 50 samples with the most significant entropy, and (iii) all samples. We processed the individual frames across each of these subsets with standard normalization and resizing owing to their resemblance of characteristics to that of image data. Dimensions after resizing the frames were set as 224x224 for training on all models used in this study, except InceptionV3 which requires input dimensions of 299x299.

\subsection{Deep Learning Models}
The premise of leveraging deep learning (DL) models toward a working solution to this problem is because of the scope of the state-of-the-art CNN models trained on image datasets. These can demonstrate a promising framework that can be deployed in clinical medicine. In this study, we model the dataset using five benchmark CNNs, each having its peculiar nature of intuition in the algorithm. Increasing the depth of the neural networks increases the parameter space and complexity. A wide range of generalization failures of DL models is attributed to the high density of the connections. In this study, we constrained the dimensionality of the models by taking relatively shallower versions of the respective models. The selection of pretrained models (Imagenet weights) was done empirically.

\subsubsection{RESNET 18}
Resnet \cite{he2016deep} is a popular and widely used DL network for identifying images. The premise upholding the generalization capability in Resnet is due to its concept of trap-door connections across the depth of the network. Due to the motivation of enabling time-sensitive querying to the model, the ResNet-18 network was used in this study.

\subsubsection{INCEPTION V3}
GoogLeNet primarily precedes the structure of inception. Inception's fundamental feature is that it pulls information from different scales of an image using several convolutional kernels before concatenating it to produce a better representation of the image. \cite{szegedy2015going}.
Two key aspects set Inception-v2 apart from Inception-v1. The 5x5 convolution is split into two 3x3 convolutions as the initial step. The second method involves splitting up a nxn convolutional kernel into two smaller convolutions, 1xn and nx1. This methodology helps reduce parameter space without any significant loss in performance. Inception-v3 specifically employs Batch Normalization \cite{saeedi2020novel} that helps in better generalization.

\subsubsection{Squeezenet V1}
The presence of a fire module (consisting of two parts viz., the squeeze part and the expand part), is at the heart of SqueezeNet. The squeeze component consists of a 1x1 convolutional kernel and a 1x1 convolutional layer. The expanded part consists of one convolutional kernel and three convolutional layers. The 1x1 and 3x3 feature maps are concatenated in the expanded layer. A comparison of the ImageNet dataset reveals that SqueezeNet and AlexNet have roughly equal accuracy \cite{iandola2016squeezenet} despite an enormous difference in trainable parameters. In this study, we model the data using Squeezenet V1.

\subsubsection{Mobilenet V1}
The depthwise separable convolution is the fundamental unit of MobileNet, which may be further broken into two smaller processes \cite{howard2017mobilenets}, pointwise convolution and depthwise convolution. The pointwise convolution is similar to the regular convolution, except that it uses 1x1 convolution kernels. Depthwise convolution, as opposed to one convolution kernel for each input channel, is a depth-level procedure \cite{howard2017mobilenets}.  This study takes into account Mobilenet V1.

\subsubsection{VGG 11 (BN)}
The VGG \cite{simonyan2015very} model has been widely used for image classification since it placed first runner-up in the ImageNet competition in 2014. The VGG architecture consists of a number convolutional layers triggered by ReLU (rectified linear unit), with a kernel size of 3x3 for the VGG convolutional layers. VGG-11, VGG-16, and VGG-19 are three VGG model versions with similar model structures. They are made up of convolutional, and pooling layers that are followed by three fully connected layers \cite{simonyan2015very}. They only differ in the number of convolutional layers (11, 16, or 19), as indicated by their names. Vanilla VGG (V19) comparably has a larger number of trainable. So, we cannot leverage VGG 11 (Batch Norm version) in this study. This helps to limit the training parameters and preserve the model performance.

\subsection{Ensemble Stacking}
Ensemble techniques are methods for producing a single optimal predictive model by combining multiple learning algorithms. The model produced outperforms the base learners on their own. Other applications of ensemble learning include feature selection, data fusion, and so on.
Stacked generalization \cite{muneeb9660782} involves stacking the results of individual estimators and using a classifier to compute the final prediction. Stacking allows capitalizing on the strengths of each individual estimator by feeding their output into a final estimator. The individual predictions of the top-3 CNN architectures are used to train the final classifier to consider the parametric weight of the contributing CNN. The normalized (softmax-based) output of each CNN is fed to the single layer perceptron model, which maps the prediction of these CNNs to optimal weights that converge during the training process. This gives a score to the contribution of each CNN. Finally, the output of the stacking model is again normalized to predict the class type. The results of querying the test data on the stacking ensemble model are shown in Fig. \ref{fig:ensembleresults}(a) and Table \ref{tab:results}. Based on the results in Table \ref{tab:results} the best three models (according to to recall scores) are stacked i.e. MobilenetV2, VGG11 (BN), SqueezenetV1. W

\subsection{majority voting ensemble}
\begin{figure*}[hbt!]
    \centering
    \includegraphics [scale=0.6]{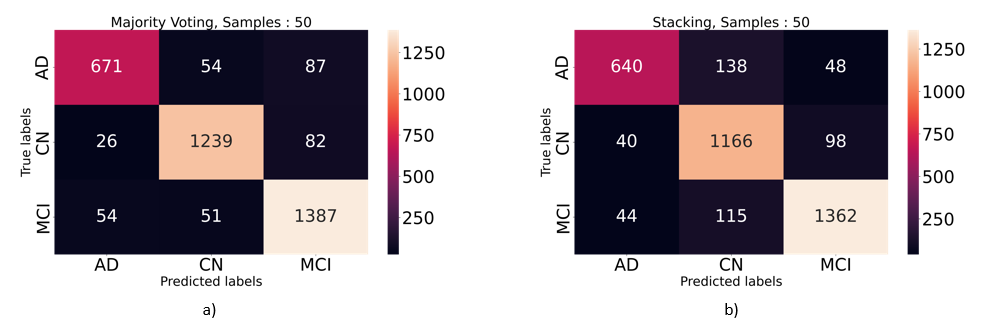}
    \caption{Ensemble Stacking and Majority Voting.}
    \label{fig:ensembleresults}
\end{figure*}

A majority voting ensemble is a machine learning ensemble model that combines predictions from several other models using the statistical mode operation. The majority voting preserves the confidence of the CNNs by treating each of the votes equally. In this experiment, majority voting yielded a tie for approximately 2\% of the test data. In this scenario, the test data was predicted as \emph{CN} class (owing to its larger distribution in the training data). Fig. \ref{fig:roc} and Fig. \ref{fig:ensembleresults}(b) illustrate the performance of the majority voting approach on the top-3 CNNs viz., VGG11(BN), SqueezenetV1 and MobilenetV2 trained on 50 high entropy samples of each scan in the dataset. The top-3 architectures were chosen because of their recall scores when trained on the sampled dataset (as shown in Table \ref{tab:results}.  

\section{Results}
\begin{figure*}[hbt!]
    \centering
    \includegraphics [width=1.05\linewidth]{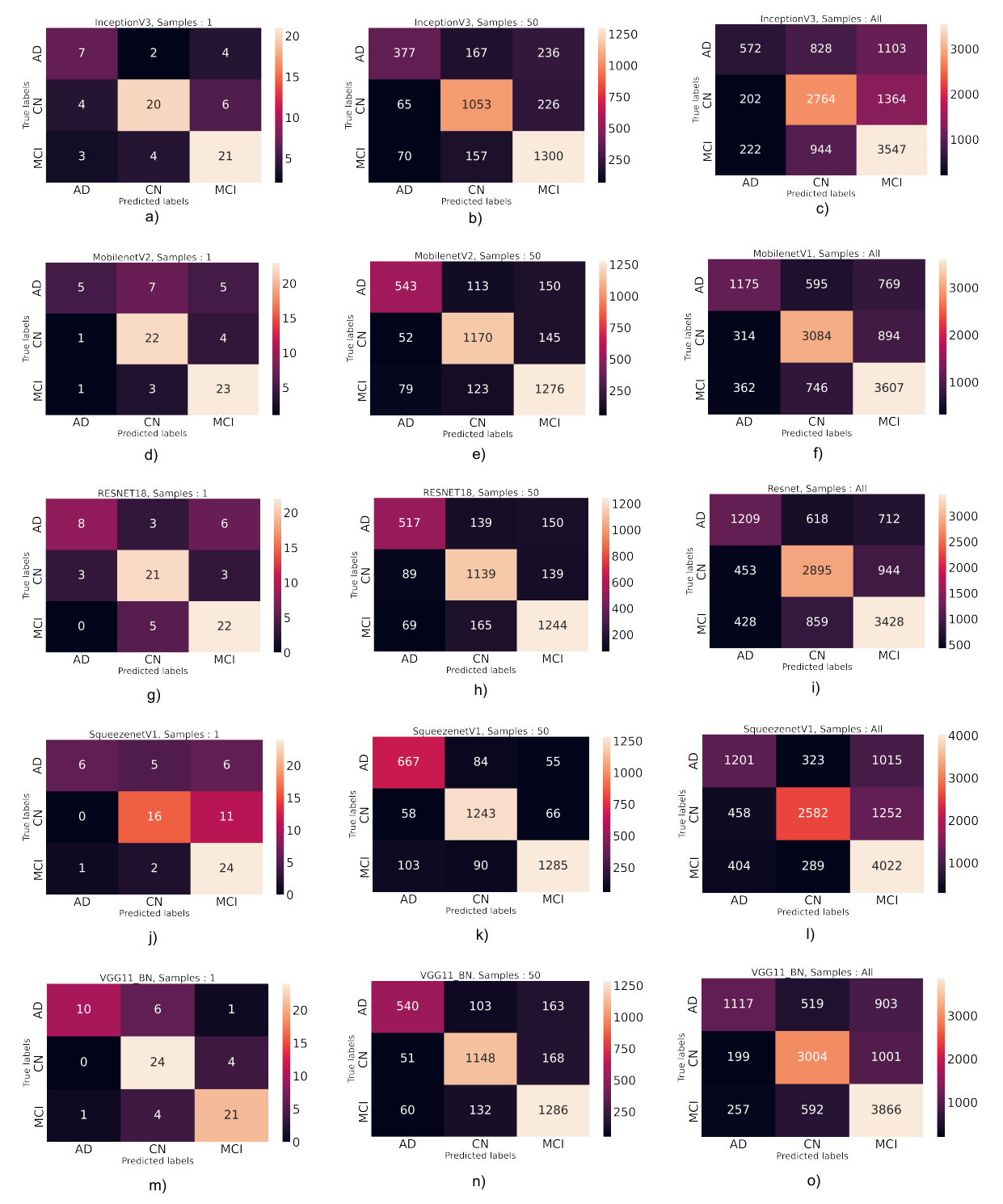}
    \caption{Illustrated Confusion Matrices of test data evaluated on different models (\textit{row wise}) and sample sizes (\textit{column wise}).}
    \label{fig:testresultsCNNs}
\end{figure*}

\subsection{Experimental Setup}
A machine with Intel i7 core 11th generation processor, 64GB RAM and 8GB Nvidia Quadro P4000 graphics card (running Ubuntu 20.04, Python 3.7, and Pytorch v1.8.1cu11) was used for conducting the experiments in this study. 

\subsection{Performance Metrics}
The data has been split into 75\% for training and 25\% for testing. The confusion matrix compares actual and predicted values to find mislabelled or incorrectly anticipated values. True Positive (TP), True Negative (TN), False Positive (FP), and False Negative (FN) are the four components (FN).All the performance metrics have been calculated using confusion matrices of each classifier, as illustrated in Fig. \ref{fig:testresultsCNNs}. Table \ref{tab:results} shows the performance of the individual models across different data samples and metrics, as defined below:

Precision: The ratio of properly expected positive output to the overall number of positive results expected.

\begin{equation}
    Precision=\frac{TP}{TP+FP}
\end{equation}

Recall (Sensitivity): The ratio of accurately predicted positive outputs to the actual positive outcomes in the class.

\begin{equation}
    Recall=\frac{TP}{TP+FN}
\end{equation}
                                                 
F1 Score: Precision and Recall were weighted, and the result was calculated. Therefore, this score includes both false positives and false negatives.

  \begin{equation}
    F1 Score=\frac{Recall \times Precision}{Recall+Precision}
\end{equation}

Specificity: Concerned with how well it forecasts the actual negative class.
  \begin{equation}
    Specificity=\frac{TN}{TN+FP}
\end{equation}             

Accuracy: The proportion of correctly predicted observations to all predicted observations.

\begin{equation}
    Accuracy=\frac{TP+TN}{Total}
\end{equation}

\begin{table}[htb]
    \centering
    \caption{Parameters set while training the models.}
    \begin{tabular}{lp{0.3\textwidth}}
    \hline \hline
\textbf{Hyperparameter}     &  \textbf{Space} \\ \hline
       Epochs  & 50 \\
       Learning Rate & 0.001 \\
       Loss &  Categorical Crossentropy \\
       Optimizer  &  SGD \\
       Batch Size  &  2 (for sampling 1 sample), 16 (for 50 samples and ALL samples) \\ \hline \hline
    \end{tabular} 
    \label{tab:parameterstable}
\end{table}

\begin{table*}[]
\caption{Performance metrics of the models.}
\centering  
\begin{tabular}{lcccc} \hline 
\textbf{Model}             & \textbf{Sample} & \textbf{Avg. Precision} & \textbf{Avg. Recall} & \textbf{Val Accuracy} \\ \hline
\multirow{3}{*}{\textbf{INCEPTION V3}} &  1       &             0.68&           0.68&            0.68\\
                    &  50       &             0.75&           0.75&            0.75\\
                    &  ALL       &             0.59&           0.60&            0.60\\ \hline
\multirow{3}{*}{\textbf{Mobilenet V2}}  &1         &             0.71&           0.70&            0.70\\
                    &         50&             0.82&           0.82&            0.82\\
                    &      ALL   &             0.68&           0.68&            0.68\\ \hline

\multirow{3}{*}{\textbf{Resnet 18}}  & 1  &     0.72    &        0.72     &     0.71                 \\
                   & 50 &        0.79 &    0.79         &           0.79          \\
                    & ALL        &             0.65&           0.65&            0.65\\ \hline
\multirow{3}{*}{\textbf{Squeezenet V1}}  &  1       &             0.69&           0.65&            0.65\\
                    &         50&             \textbf{0.88}&           \textbf{0.88}&            \textbf{0.88}\\ 
                    &    ALL     &             0.69&           0.68&            0.68\\ \hline
\multirow{3}{*}{\textbf{VGG 11 (BN)}}     &      1   &             \textbf{0.79}&           \textbf{0.77}&            \textbf{0.77}\\ 
                    &         50&             0.82&           0.82&            0.82\\
                    &ALL         &             \textbf{0.70}&           \textbf{0.69}& \textbf{0.69}\\ \hline 

\multirow{3}{*}{\textbf{Majority Voting Ensemble}}               &         50 &             \textbf{0.90} &           \textbf{0.89} &            \textbf{0.90}\\ \hline  
\end{tabular} 
\label{tab:results}
\end{table*}

\begin{figure*}[hbt!]
    \centering
    \includegraphics [scale=0.75]{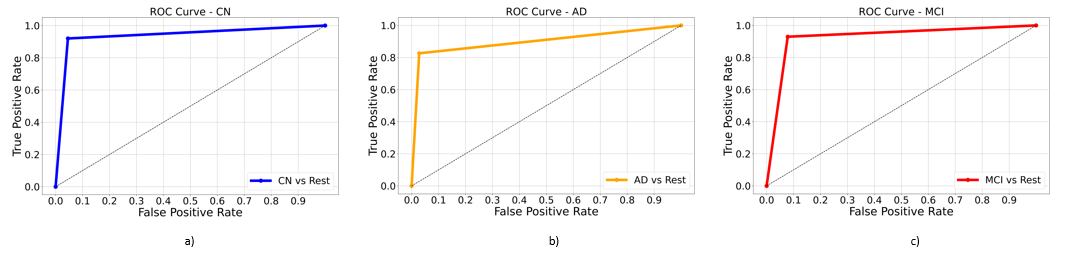}
    \caption{ROC curves for majority voting algorithm on  MobilenetV2, VGG11 (BN) and SqueezenetV1 across different classes (OneVsAll).}
    \label{fig:roc}
\end{figure*}

\section{Discussion}
\subsection{Results}
The results of our experiments signify that the performance of training on the sample category having 50 samples with the most significant entropy is better than the other two sample types in almost all the models. The reason for this result is intriguing. A lot of all-black and/or redundant frames exist within the recording instances due to the latency in switching the recording machine on/off. This entails that most of the information within MRI recording is contained in a limited subset of the frames. Most of the MRI data is redundant, so it is eliminated prior to modeling, and the most notable samples are chosen by considering the sample's entropy. CNNs learn features owing to the spatial variability in the data. Entropy is maximum when enough information and variability are contained in the data. This establishes that entropy as a measure of data quality is cohesive to the modelling approach used. This inference can be used for real-time analysis of such a system in clinical practice. It could fasten early diagnosis of Alzheimer's disease that is based on prediction only on certain parts of the scan that are essentially rich in information content.
\indent Another intriguing result deduced from the experiments suggests that the models with low parameter space outperform others. This is generally true when the dataset size is limited. However, this could be attributed to the complex nature of the deeper models. The depth of the models introduces high variance in training data and low performance in the test data. In this problem, the data essentially is grayscale. Training grayscale images on complex deep models that are pretrained on color images, introduces poor generalization. This reason also extends to the poor performance of the stacking ensemble method. Stacking adds to the complexity of the model and overfits the data concerning the problem at hand.
\indent A slight difference in the population distribution of training data has some slight effect on the ROC of each class (as shown in Fig. \ref{fig:roc}). However, the proposed model shows approximately equal numbers across the classes in terms of recall.
\indent Observing a higher recall is the sought objective to characterize a better performance in this type of a problem. We observe that majority voting outperforms the individual modelling approaches. The bias and variance terms are compositions in error. A significant portion of error could also occur due to the variance in the predictions of the individual models. In this case, the majority voting technique tends to reduce the prediction variance by taking the support of 2 out of 3 individual models. The performance metrics of the majority voting ensemble are shown in Table \ref{tab:results}. In this approach, observing a higher recall and precision could directly result in a decrease in the error. 
\subsection{COMPARISON WITH OTHER STUDIES}
Termenon et al. \cite{termenon2012two} describe a two-stage sequential ensemble wherein samples are assessed by a 2nd phase classifier where the outcome by the first classifier lies inside a low confidence output interval (LCOI) and employ GM density maps to develop feature vectors. The training process is divided into three steps: training the first classifier, determining the first classifier's LCOI, and training the second classifier using the collected data whose result lies inside the LCOI. The obtained results are better than those pertaining to such a database based on relevance vector machine (RVM). 
Sorensen et al. \cite{sorensen2016early} finding point to the presence of aberrant hippocampus texture in MCI, as well as the notion that texture could be used as a predictive neuroimaging indicator for earlier cognitive deficits. These techniques are dependent on a prior understanding of the scope and distribution of AD. T1-weighted MRI scans from the ADNI collection were used to train and validate the texture markers. Tang et al. \cite{tang2015baseline} present a new predictor for the progression of moderate cognitive impairment (MCI) to Alzheimer's disease (AD) based on biologically selected features. The shape diffeomorphometry patterns of subcortical and ventricular components are used to create this prediction. The best predictor is a collection of the form diffeomorphometry sequences of the right caudate, right thalamus, left lateral ventricle, left hippocampus, and bilateral putamen produced by a feature selection technique.
 Li et al. \cite{li2018alzheimer} present a classification approach for Alzheimer's disease diagnosis based on DenseNets and MR images. Without domain expert knowledge, the approach is a data-driven method that involves concurrently learning features and categorization and preprocessing MR images that do not necessitate rigorous registration and segmentation. The research of Liu et al. \cite{liu2020new} provides a new method for detecting Alzheimer's disease based on spectrogram characteristics extracted from speech signals. This strategy can help family members understand their loved one's illness progression early, allowing them to take preventive action and among the models examined, the LogisticgressionCV performed the best. Battineni et al. \cite{battineni2021deep} used deep learning-based convolution neural networks (CNN) to accurately classify the several phases of dementia photos, achieving an accuracy of 83.3 percent, which is greater than other standard classifiers such as support vectors and logistic regression.
 \indent Our work focuses on assessing benchmark Deep Learning models' capabilities for categorizing MRI data. Our findings demonstrate that training on the sample category with 50 samples having the highest significant entropy yields better results than the other two sample kinds and it is 88 percent in case of SqueezenetV1 model.

\subsection{Tradeoffs of Performance Metrics}
The model's parameters and hyperparameters can be changed to improve precision. While adjusting, it is possible to notice that higher precision generally results in a lower recall, and higher recall results in lower accuracy.
Similarly, any machine learning model's recall value can be altered by adjusting multiple parameters or hyperparameters. For any model, a higher or lower recall has a specific meaning: A high recall indicates that the majority of positive instances (TP+FN) will be identified (TP).
This leads to a greater number of FP measurements and a decrease in overall accuracy. However, suppose the outcome is low recall. In that case, it indicates that there were many FNs (should have been positive but labeled negative), which means that if the results find a positive example, it may be more confident that it is a true positive. Furthermore, while F1 is not as intuitive as accuracy, it is usually more advantageous, especially when unequal class distribution is. Accuracy improves when the cost of false positives and false negatives is the same. If the cost of false positives and false negatives is significantly different, it is better to consider both Precision and Recall.

Furthermore, sensitivity (recall) and specificity are inversely proportional. Susceptible tests produce more positive results for patients with illness, whereas precise tests reveal no illness in patients who do not have a finding. To provide a complete diagnosis, sensitivity and specificity should always be considered concurrently. Furthermore, when datasets are symmetric, and the values of false-positive and false-negatives are nearly similar, accuracy is a good quality measure. As a result, other parameters are also important in determining a model's performance.

\subsection{ Behaviour of Models Used}

Each VGG block is made up of layers of 2D Convolution and Max Pooling. The model's ability to fit more complex functions increases as the number of layers in CNN increases. As a result, adding more layers promises better performance. This is not to be confused with an Artificial Neural Network (ANN), where adding layers does not always improve performance. Backpropagation is a neural network weight update algorithm that makes minor changes to each weight in order to reduce the model's loss. Each weight is updated so that it moves in the direction of decreasing loss. This is simply the weight's gradient as determined by the chain rule. However, as the gradient returns to the initial layers, the value increases with each local gradient. As a result, the gradient shrinks and shrinks, resulting in minor changes to the initial layers. As a result, training time is greatly increased. The problem is solved if the local gradient equals one.
This is where ResNet comes in because it achieves this through the identity function. As a result, the gradient's value does not decrease as it is back-propagated because the local gradient is 1. Another type of 50-layer deep convolutional neural network architecture is deep residual networks (ResNets), such as the popular ResNet-50 model (CNN). By inserting shortcut connections, a residual neural network converts a plain network into its residual network counterpart. Because they have fewer filters, ResNets are less complex than VGGNets. ResNet does not allow the vanishing gradient problem. The skip connections act as gradient superhighways, allowing the gradient to flow freely. This is also one of the primary reasons why ResNet is available in several versions, including ResNet50, ResNet101, and ResNet152.

Inception was created to reduce the computational load of deep neural nets while maintaining cutting-edge performance. The creators of inception were looking for a method to scale up neural networks without raising the computational cost because computational efficiency decreases as the network grows deeper. ResNet is concerned with computational accuracy, whereas inception is concerned with computational cost. Deeper networks should outperform shallower networks in theory, but deeper networks outperform shallower networks in practice due to an optimization problem rather than overfitting. In short, the more difficult it is to optimize, the deeper the network. To attain more accuracy, computer vision networks are getting deeper and more complex. On the other hand, deeper networks give up size and speed. The object identification task must be able to be completed on a platform with limited processing resources in real-world applications like autonomous automobiles or robotic visions. To address this issue, MobileNet, a network for embedded vision applications and mobile devices, was created. The aim of MobileNet is to create lighter deep neural networks by using depthwise separable convolutions. The convolution kernel or filter is applied to each channel of the input image by weighting the input pixels with the filter before moving on to the next input pixels across the pictures. This regular convolution is used only in the first layer of MobileNet.The following layers combine depthwise and pointwise convolutions to create depthwise separable convolutions. Depthwise convolution is used to individually convolve each channel. The output image will also have three channels if the input image has three. The input channels are filtered by means of this depthwise convolution. Following that is pointwise convolution, which is similar to regular convolution but contains a 1x1 filter. The purpose of pointwise convolution is to combine the depthwise convolution output channels to create additional features. As a result, the computational work required is less than that required by traditional convolutional networks. Other cutting-edge convolutional neural networks, such as VGG16, VGG19, ResNet50, InceptionV3, and Xception, outperform MobileNet. MobileNets are thin deep neural networks designed for mobile and embedded vision applications. It employs depthwise separable convolutions in a streamlined architecture and two simple global hyperparameters to efficiently trade off accuracy versus latency. MobileNet could be used for various applications such as object detection, fine-grain classification, face recognition, large-scale geolocation, and so on. The advantages of using MobileNet over other cutting-edge deep learning models are as follows. It reduced the network size to 17MB and the number of parameters to 4.2 million. It is faster and more suitable for mobile applications. It has a low-latency convolutional neural network.
There are always drawbacks to advantages, and with MobileNet, it's the accuracy. Despite being smaller, having fewer parameters, and performing faster, MobileNet is less accurate than other cutting-edge networks. ResNet models reduce training time while increasing accuracy by not activating all neurons in every epoch. Furthermore, the model employs an ingenious strategy for improving model training performance by learning a feature once and then not attempting to learn it again; instead, it focuses on learning additional features. By introducing pretrained models and increasing model depth, VGG significantly improved speed and accuracy. The nonlinearity of the model increased as the number of layers with smaller kernels increased. Unlike Inception v1–v3, the Inception-ResNet-v2 model employs residual networks to improve the original model's accuracy and convergence speed.

\subsection{Behaviour of ensemble stacking}
The best model should be chosen from among several alternatives. In general, ensemble models produce lower variance and bias while increasing accuracy. Furthermore, increasing the size of the ensemble improves test results. As a result, ensembles frequently triumph in competitions. Each method has distinct characteristics. On the other hand, model ensembles are not always ideal for various reasons. It takes longer to train than an individual model because many base models must be assembled. The poor test results of the stacking ensemble could be attributed to high variance in the stacking-based ensemble model, leading to overfitting. Furthermore, it may not be advantageous in the case of a small dataset. 

\subsection{Future Scope}
This research can benefit all researchers working in this field around the world. This study's future scope cannot be limited to hardware implementation. Applications can be expanded to include other types of medical data such as images, signals, and so on. Furthermore, additional classifiers, neural networks, and other AI techniques can be applied to the same or different datasets to improve Alzheimer's detection.

\section{Conclusion}
This work uses deep learning techniques to categorize Magnetic Resonance Imaging (MRI) data for Alzheimer's disease (AD) on the ADNI 3Yr 3T dataset. Deep learning outperforms classical machine learning in terms of finding detailed structures in complicated, high-dimensional data, especially in the area of healthcare. The ultimate goal of this study is to utilize the efficiency of five different models viz., Resnet18, SqueezenetV1, VGG11(BN), InceptionV3 and MobilenetV2, for achieving high recall and accuracy. The MRI scans in the dataset procured from ADNI are volume-data. Each scan in the dataset is split into images of various samples, and the most significant samples are chosen by considering the sample's entropy. Our experiment findings show that training on the sample category with 50 samples that have the most significant entropy performs better than the other two sample classes, and majority voting works better than the other modeling strategies as the majority voting method tends to minimize the variance in the predictions. We report a recall score of 0.89 on the test data with a test accuracy of 90\%.
\bibliographystyle{IEEEtran}
\bibliography{main}
\EOD
\end{document}